\PassOptionsToPackage{table,dvipsnames}{xcolor}
\documentclass{applemlr}

\usepackage{amsmath}
\usepackage{amssymb}
\usepackage{amsfonts}
\usepackage{amsthm}
\usepackage{colortbl}
\usepackage{tabularx}
\usepackage{booktabs}
\usepackage{enumitem}
\usepackage{graphicx}
\usepackage{url}
\usepackage{xurl}

\usepackage{amsmath,amsfonts,bm}

\def\eqref#1{equation~\ref{#1}}

\def\1{\bm{1}}

\DeclareMathAlphabet{\mathsfit}{\encodingdefault}{\sfdefault}{m}{sl}
\SetMathAlphabet{\mathsfit}{bold}{\encodingdefault}{\sfdefault}{bx}{n}

\newcommand{\E}{\mathbb{E}}

\newcommand{\R}{\mathbb{R}}

\newcommand{\Var}{\mathrm{Var}}

\makeatletter
\AtBeginDocument{\urlstyle{sf}}
\makeatother

\usepackage{placeins}
\usepackage{xcolor}

\newcommand{\sphere}{\mathbb S}

\newcommand{\Bet}{\mathrm{Beta}}
\newcommand{\Ber}{\mathrm{Bernoulli}}
\newcommand{\eprd}{e_{t\mid t-1}}
\newcommand{\nprd}{n_{t\mid t-1}}
\newcommand{\epst}{e_{t\mid t}}
\newcommand{\npst}{n_{t\mid t}}
\newcommand{\Rset}{\mathcal R}

\DeclareMathOperator{\logit}{logit}

\title{BRAG: Bayesian Retrieval Augmented Generation}

\author{Javier Movellan}
\author{Manjot Bilkhu}
\author{Young-mok Jung}
\author{Sirajul Salekin}
\author{Henry Tran}
\author{Zhao Huang}

\affiliation{Apple}

\abstract{
Retrieval-augmented generation (RAG) ranks memory chunks by semantic
similarity to the query. Many methods improve on this by learning from
feedback on what was retrieved, but each is a separate heuristic, with
no account of when it is the right choice. We introduce BRAG, a
generative model of retrieval with use. Each chunk generates the queries
it answers from a mixture of von Mises--Fisher components anchored at
its original embedding, and each chunk has a latent utility that drifts
over time. The model gives the optimal inference, a closed-form filter
that stores a few keys and two scalars per chunk and updates them online
from the labels of a curator, and it frames retrieval as a control
problem. It also shows when existing methods are optimal. Query
averaging, Rocchio-style feedback and frequency reranking are special
cases with fixed parameters, NUDGE is the case of one key per chunk with
no dynamics, and LADER the limit of one key per past query. On two
public long-term memory benchmarks, BRAG matches these methods on
queries asked before, as expected, because when nothing changes over time
the assumptions behind these special cases hold. On harder problems, where a new
chunk makes an old one obsolete, BRAG outscores them. They all stay at
plain-search level (Recall@5 0.67--0.71), while BRAG retrieves the new
chunk 0.80--0.86 of the time. We also ran BRAG on three of NUDGE's
seven text benchmarks, and it matches or beats NUDGE on all three. On nine months of a personal agent's
memory (20{,}640 chunks), BRAG raises Recall@5 on held-out queries from
0.75 to 0.98 and Recall@1 from 0.44 to 0.97, with little change on
chunks never queried.
}

\date{\sffamily\today}

\begin{document}

\maketitle

\section{Introduction}\label{sec:context}

AI agents operating over months or years accumulate ever-growing
collections of conversations, documents, code, and other
artifacts. Retrieval-augmented generation (RAG) systems
\citep{lewis2020} index these artifacts by semantic similarity, so a
query retrieves the chunks whose embeddings lie closest to it. But
similarity is not what matters to the consumer of the retrieved
content. What matters is whether a chunk holds the information the
query asks for, and that changes with use. Some chunks are asked
about again and again, and others go stale when newer chunks replace
them.

Libraries solved an analogous problem with two independent organizing
systems. \emph{Access priority} migrates books toward or away
from the front shelf according to anticipated demand, and
\emph{semantic organization} shelves related books
together. Bayesian models of borrowing history have long been used to
predict the first \citep{burrell1985}, and the same class of models
has been shown to explain classical phenomena of human memory such as
forgetting curves, learning curves and spacing effects
\citep{anderson1991}. Building on this tradition, we introduce BRAG, a
generative model of retrieval with use. Each query is generated by one
chunk, the one holding the information it asks for. Each chunk has a
time-dependent latent \emph{utility}, the probability that the query
at time $t$ is generated by it. The per-chunk query distribution is
also a latent variable, parameterized by a mixture of
von Mises--Fisher components with a prior centered on the chunk's
original embedding. Beliefs over utility are Beta distributions, updated online
as evidence accumulates and relaxed toward a low-utility,
high-uncertainty equilibrium in the absence of new evidence.

The evidence comes from labels. Each query retrieves a set of chunks,
and a label says which of them, if any, contains the information asked
for. The label can come from the user's behavior after the retrieval or
from a curator LLM that checks the retrieved chunks. In BRAG a chunk's
semantic embedding is not a fixed address but a parameter of the
distribution of queries the chunk generates, and that distribution is
learned from use. The resulting filter has small memory and
computational cost per chunk and is linear in corpus size, so it scales
to production RAG systems. Retrieval is formulated as an optimal control problem given the
posterior distribution over utility conditioned on the history of
retrieval and the observed query.

Many methods already learn from
this kind of feedback, such as averaging past queries into a chunk's vector,
Rocchio's relevance feedback \citep{rocchio1971}, reranking by
frequency or recency, NUDGE \citep{nudge2025}, which moves each
embedding toward the queries it answered, and LADER \citep{lader2023},
which scores documents by a kernel density over their past
queries. Each is a separate heuristic, with no account of when it is
the right choice. BRAG provides that account. Under its generative
model these methods are special cases with fixed parameters, which
says when each is optimal and lets the parameters be learned instead.

We make four contributions.
(i) A generative model of retrieval with use, in which each chunk
generates queries from a von Mises--Fisher mixture and has a utility
with Ornstein--Uhlenbeck dynamics, giving optimality conditions for both
inference and retrieval.
(ii) A closed-form recursive filter for the utility beliefs and an
online EM adaptation of each chunk's keys, at the cost of a few keys
and two scalars per chunk.
(iii) A unifying account of feedback-based retrieval, with query
averaging, Rocchio, frequency reranking, NUDGE (one key per chunk, no
dynamics) and LADER (one key per past query, no dynamics) as special
cases.
(iv) An empirical evaluation on two public long-term memory benchmarks,
on three of NUDGE's seven text benchmarks, on the TripClick search logs
that LADER was tested on, and on nine months of a personal agent's
memory. BRAG matches the special cases when nothing changes over time,
and outscores them when a new chunk makes an old one obsolete, with
Recall@5 of the new chunk 0.80--0.86 against 0.67--0.71 for every
method without dynamics.

Section~\ref{sec:space} lays out the generative model,
\S\ref{sec:infer} derives the recursive filter,
\S\ref{sec:adapt} develops the online key adaptation, and
\S\ref{sec:retrieval} describes retrieval as control.
\S\ref{sec:sim} evaluates BRAG on public benchmarks and a personal
memory corpus, \S\ref{sec:related} discusses related work, and
\S\ref{sec:discuss} concludes.

\section{The BRAG Generative Model}\label{sec:space}

Each query is generated by one chunk, the chunk that holds the
information the query asks for. We define the utility of a chunk at
time $t$ as the probability that the query at time $t$ is generated by
it. Time steps are taken small enough that at most one chunk generates
a query in a step. We represent the internal beliefs about
the utility of a chunk using a Beta distribution, in the
expectation--concentration parameterization $(e,n)$ of the standard
Beta-Binomial conjugate model \citep{gelman2013}, where $e$ is the expected
utility and $n$ can be interpreted as the effective sample size behind it.
For example $\Bet(0.8,10)$ represents the belief distribution carried
by $n=10$ observed trials, including prior pseudo-observations, with a
rate of $e=0.8$, i.e., ``eight hits in ten looks''. A large $n$
indicates that $e$ is a high precision estimate that one new
observation barely moves, and a small $n$ indicates that $e$ is only a
tentative estimate that a single observation can move substantially.

\paragraph{System Dynamics.}
Each chunk $i$ has a latent utility $X_t^i\in(0,1)$ that changes over
time, e.g., a chunk that is useful today may not be as useful in the
future.  The dynamics are set by five model parameters, $(e_0,n_0)$,
$(\bar e,\bar n)$, and $h$. The belief distribution about the utility
of a new chunk starts at a shared prior, $X_0^i\sim\Bet(e_0,n_0)$, where
$e_0$ is the assumed starting interest of new content and $n_0$ how
firmly that assumption is held (a small $n_0$ lets the first few
observations move a new chunk quickly).  In the absence of
observations, the belief relaxes toward $\Bet(\bar e,\bar
n)$. Typically $\bar e$ is small, since old chunks tend to have little
utility, and $\bar n$ is moderate, so the long-term expectation is held
with some, but not great, confidence. The half-life $h$ sets the speed
of that relaxation. The expected utility's excess over $\bar e$ halves
every $h$ days, so a large $h$ means that a chunk's utility persists
for a long time. The parameters can be set by hand or fit to data
(\S\ref{sec:sim}). These dynamics are
implemented as an Ornstein--Uhlenbeck (OU) process on the utility
logit, $L_t^i=\logit X_t^i$. Working in logit space keeps the utility
inside $(0,1)$ and turns the relaxation into a plain linear
mean-reversion, the simplest process with a half-life,
\begin{equation}\label{eq:ou}
  L_t^i=a\,L_{t-1}^i+(1-a)\,\bar\ell+\sigma Z_t^i,\qquad Z_t^i\sim\mathcal N(0,1)\ \text{i.i.d.},
\end{equation}
where $a=2^{-1/h}$,   and the equilibrium mean,
variance and noise scale are all derived from $(\bar e,\bar n)$ as
$\bar\ell=\logit\bar e$, $\bar v=1/\big(\bar e(1-\bar e)\,\bar n\big)$,
$\sigma=\sqrt{(1-a^2)\,\bar v}$.

\paragraph{Observation Model.}
Let $U_t^i \in \{0,1\}$ indicate that chunk $i$ generates the query at
time $t$. Given the state, $U_t^i \sim\Ber(X_t^i)$, independently
across chunks. When chunk $i$ generates the query, i.e., when
$U_t^i=1$, a query vector $Q_t\in\sphere$ is drawn
from a mixture of a von Mises--Fisher (vMF) distributions
\citep{banerjee2005,mardiajupp2000}. First a topic $c$ for the query
is drawn with probability $\omega^{i,c}$ and then the query vector $q$
is drawn from the vMF distribution centered at the key vector
$k^{i,c}$ for chunk $i$ topic $c$, i.e.,
\begin{align}\label{eq:emit}
  p\big(q\mid U_t{=}1_i\big)&= c(\psi)\, \phi^i(q),\;\;\;\phi^i(q) = \sum_{c=1}^{C_i} \omega^{i,c}\,\,e^{\psi\,q^{\top}k^{i,c}}
\end{align}
where $\psi>0$ is the keys' precision parameter, $c(\psi)$ a chunk-independent
normalizer and $\phi^i(q)$ is the chunk's \emph{hit weight}. Here
$U_t=(U_t^1,\ldots,U_t^N)\in\{0,1\}^N$ is the use vector across all
chunks, and $U_t=1_i$ denotes the one-hot event that chunk $i$ generated the
query. 

\section{Inference about chunk utility}\label{sec:infer}

Every chunk carries two numbers, its belief pair $(e,n)$. On each
update from time $t-1$ to time $t$, the filter does two things to them, a \emph{predict} step, which
relaxes the belief toward the equilibrium prior in proportion to the
time elapsed, and an \emph{update} step, which folds in what the
query at time $t$ revealed.  The evidence comes in two forms, the
observed query $q_t$ and an indicator $u_t^i=1$ for which of the
retrieved chunks, if any, contains the information the query asks for.  The whole recursion is summarized below.

\begin{center}\small
\setlength{\tabcolsep}{8pt}\renewcommand{\arraystretch}{1.4}
\begin{tabular}{@{}ll@{}}
\toprule
Init & $(e_{0\mid 0}^i,\,n_{0\mid 0}^i)=(e_0,\,n_0)$\\[6pt]
Predict & $\eprd^i=a\,e_{t-1\mid t-1}^i+(1-a)\,\bar e,\qquad \dfrac{1}{\nprd^i}=\dfrac{a^2}{n_{t-1\mid t-1}^i}+\dfrac{1-a^2}{\bar n}$\\[10pt]
Update & $\epst^i=\dfrac{\eprd^i\,\nprd^i+u_t^i}{\nprd^i+1},\qquad \npst^i=\nprd^i+1$\\[10pt]
\bottomrule
\end{tabular}
\end{center}
The rest of this section explains each row.

\paragraph{Predict.}
The discrete-time OU process \eqref{eq:ou} on the logit of utility
defines the dynamics of the model, while the belief lives in the Beta
family. The Laplace bridge \citep{mackay1998,hobbhahn2022}, which maps
a Beta to a logit-Gaussian and back, turns one OU step into a recursion
on the Beta parameters $(e,n)$. Once the bridge is applied, the filter
no longer needs the underlying logit process and propagates the belief
directly in the Beta family. The predict step carries
the posterior at $t{-}1$ to the prior for $t$,
\begin{equation}\label{eq:predict}
  \eprd^i=a\,e_{t-1\mid t-1}^i+(1-a)\,\bar e,\qquad
  \frac{1}{\nprd^i}=\frac{a^2}{n_{t-1\mid t-1}^i}+\frac{1-a^2}{\bar n}.
\end{equation}
The mean decays geometrically toward $\bar e$ with retention $a$, and the
inverse sample size $1/n$ blends toward the equilibrium $1/\bar n$ at rate $a^2$.
A chunk that is never queried therefore drifts to a low-utility,
moderate-confidence resting state.

\paragraph{Update.}
When a query is received $\kappa$ chunks are retrieved and, when
available, the model receives an indication of which of the retrieved
chunks, if any, contains the information asked for. The label is one Bernoulli observation of the chunk's
utility, and the update is exact Beta--Bernoulli conjugacy. The mean
moves toward $u_t^i$ and the effective sample size supporting the mean
grows by one. Labels are available for the retrieved chunks only, so
labeled chunks take this step and unlabeled chunks receive only the
predict step. 

\paragraph{Responsibility.}
Given the history of observations $h_{1:t-1}$ and a query $q_t$, the
model provides the posterior probability that chunk $i$ generated the
query, its \emph{responsibility}, a normalized, odds-weighted
softmax over the corpus,
\begin{equation}\label{eq:resp}
  \boxed{\ \gamma_t^i:=P\big(U_t^i=1\mid q_t,\,h_{1:t-1}\big)
  =\E\big[U_t^i\mid q_t,\,h_{1:t-1}\big]
  =\frac{r_t^i\,\phi_t^i}{\displaystyle\sum_{j}r_t^j\,\phi_t^j},\,
  \qquad r_t^i=\frac{\eprd^i}{1-\eprd^i}}
\end{equation}
with $\phi_t^i=\phi^i(q_t)$ the hit weight of \eqref{eq:emit}. A chunk
must be both likely useful and semantically aligned to earn a high
$\gamma_t^i$. This is the quantity the retrieval policy ranks by
(\S\ref{sec:retrieval}).

\section{Inference about the chunk query distribution}\label{sec:adapt}
For each chunk BRAG keeps a model of the distribution of queries the
chunk generates.  The parameters of the  model,
the keys $k^{i,c}$ are initialized using standard semantic encoders,
representing the prior usage distribution and are adapted based on the
queries a chunk is confirmed to respond to. In practice we use
supervised EM \citep{banerjee2005} to maximize the posterior of the
parameters given the history of observations. 

\paragraph{E-step.} When chunk $i$ is labeled as containing the
information asked for, which of its keys generated
the query is still latent, and its posterior is the within-chunk responsibility
\begin{equation}\label{eq:zeta}
  \zeta_t^{i,c}=P\big(G_t=c\mid U_t=1_i,\,q_t\big)
  =\frac{\omega^{i,c}\,e^{\psi\,q_t^{\top}k^{i,c}}}
        {\displaystyle\sum_{c'}\omega^{i,c'}\,e^{\psi\,q_t^{\top}k^{i,c'}}}.
\end{equation}
Credit is two-level. The corpus-wide responsibility $\gamma_t^i$
\eqref{eq:resp} decides \emph{which chunk}, and $\zeta_t^{i,c}$ decides
\emph{which of its keys}.

\paragraph{Batch EM rotation.} At each adaptation step we collect all
labeled queries for chunk $i$ up to the current time and run standard EM on
the vMF mixture. The E-step assigns within-chunk responsibilities $\zeta$
via \eqref{eq:zeta}. The M-step computes the responsibility-weighted
resultant for each key component and normalizes to the sphere,
\begin{equation}\label{eq:keybatch}
  k_t^{i,c}=\mathrm{normalize}\!\Big(\zeta_0^{i,c}\,k_0^i+\sum_{s\le t}\zeta_s^{i,c}\,q_s\Big),
  \qquad
  \omega_t^{i,c}=\frac{\zeta_0^{i,c}+\sum_{s\le t}\zeta_s^{i,c}}
                       {\sum_{c'}(\zeta_0^{i,c'}+\sum_{s\le t}\zeta_s^{i,c'})},
\end{equation}
where the sums run over all labeled events for chunk $i$ and
$\zeta_0^{i,c}$ is the total responsibility assigned to $n_p$ copies of the
base embedding $k_0^i$ included as a prior ($n_p{=}1$ in our experiments),
anchoring the keys near their initial position when labeled data is sparse. This is batch
EM for the vMF mixture \citep{banerjee2005}, which is standard and stable given the
accumulated data. Streaming variants using an exponential moving average on
the sphere are also possible.

\section{Retrieval as a Control Problem}\label{sec:retrieval}
The filter of \S\ref{sec:infer} and the key adaptation of
\S\ref{sec:adapt} are state estimators. They track beliefs and
geometry but do not decide what to surface, which is a control problem.  Here
we examine control policies that consume only the filter's outputs,
the belief state $\{(\eprd^i,\nprd^i)\}$ and the hit weights
$\phi_t^i$.

\paragraph{The retrieval problem.} At each period a query $q_t$
may arrive and the system must choose a \emph{retrieved set}
$\Rset_t\subset\{1,\ldots,N\}$ of at most $\kappa$ chunk indices.  The
goal is to predict which chunk will be used. Given $q_t$ and the
observed history $h_{1:t-1}$, the controller needs to guess
$U_t$. With a discount factor $\beta\in[0,1)$, the objective is
\begin{equation}\label{eq:control}
  J(\pi)=(1-\beta)\,\E\bigg[\sum_{t\ge1}\beta^{\,t-1}
          \sum_{i\in\Rset_t}\!\E\big[U_t^i\mid q_t,\,h_{1:t-1}\big]\ \bigg],
\end{equation}
where $\Rset_t=\pi(q_t,h_{1:t-1})$ is the policy's choice. The inner
expectation is the responsibility $\gamma_t^i$ of \eqref{eq:resp}, so
the objective simplifies to
\begin{equation}\label{eq:towercontrol}
  J(\pi)=(1-\beta)\,\sum_{t\ge1}\beta^{\,t-1}
     \sum_{i\in\Rset_t}\!\gamma_t^i\,,
\end{equation}
where for readability we suppress the dependence of $\gamma_t^i$ on
$q_t$ and $h_{1:t-1}$.

\paragraph{Case 1 ($\beta\to 0$).} When only the current
period matters, maximizing \eqref{eq:control} reduces to maximizing
\eqref{eq:towercontrol} term by term, which means returning the $\kappa$ chunks of
largest $\gamma_t^i$, and since the normalization term is constant for
all chunks, this is equivalent to returning the chunks with
largest $r_t^i \phi_t^i$. At $\kappa=1$ this is pure prediction, pick
the single chunk most likely to have generated the query. At $\kappa>1$, pick
the top $\kappa$. In practice this can be done in a two step process
in standard RAG databases. First we retrieve the top $m\kappa$ samples
based on the query/key semantic alignment $\phi_t^i$. These are
multiplied by the posterior odds $r_t^i$, reranked, and the top $\kappa$
are returned.  A chunk that is both semantically aligned with the query
(high $\phi^i_t$) and has demonstrated historical utility (high $r^i_t$) is
promoted, while a chunk with high alignment but low posterior belief, or
vice versa, may be overtaken. This is the policy used in all
experiments in this paper.

\paragraph{Case 2 ($\beta \to 1$).} For $\beta>0$ the controller
optimizes over the long run and must balance exploitation against
exploration, since a chunk with low expected utility but high uncertainty may
be worth surfacing in order to learn about it. As $\beta\to1$, posterior
sampling \citep{thompson1933,russo2018} is the standard approximation.

\paragraph{Case 3 ($0<\beta<1$).} For intermediate discount factors,
approximately optimal control methods such as reinforcement learning
can learn the retrieval policy, with the filter's belief state and hit
weights as the state. We evaluate only the greedy case (Case 1) here.

\section{Experiments}\label{sec:sim}

We evaluate BRAG in four settings, retrieval of information
accumulated between a user and a personal AI agent
(\S\ref{sec:exp-personal}), two public long-term memory benchmarks
with many users that include scenarios in which new information makes
old information obsolete (\S\ref{sec:exp-public}), three of NUDGE's
seven text benchmarks run with its code and splits
(\S\ref{sec:exp-nudge}), and the benchmark of real search logs that LADER was tested on
(\S\ref{sec:exp-lader}).

\subsection{Personal memory}\label{sec:exp-personal}

\paragraph{Corpus and queries.}
The corpus is a RAG database of 20,640 chunks drawn from 2,742 markdown
documents accumulated over 9 months of daily interaction between a
user and a personal AI agent. The agent reviewed the transcript of the
most recent ten weeks and selected 289 queries representative of common
information needs, each with its correct target chunk. For every query
it then wrote four additional queries targeting the same chunk. Because
a chunk can be relevant to more than one topic, the five queries for a
chunk are not only rewordings but may be distinct ways of asking about
different topics.

\paragraph{Protocol.}
For each chunk, one of its five queries is held out and never used for
training. The remaining 1,156 queries are shuffled and spread over 44
simulated days. Each day the system retrieves chunks for that day's
queries, and the curator labels them. In all our experiments only positive
labels are applied. A chunk marked as containing the information asked
for takes the update step, and every other chunk takes only the predict
step. A label is applied only if the
labeled chunk is within the top 100 retrieved by that condition at the
time of the query. This matches how the curator works in our deployed
system. When no chunk among the top $\kappa$ retrieved holds the
answer, it widens the search up to the top 100 before giving up, so a
chunk outside the top 100 can never be labeled.  At the end of each
day the labels are applied and the keys and beliefs updated
(\S\ref{sec:infer}--\S\ref{sec:adapt}). After day 44 we score the 281
held-out queries that have a correct chunk. Each condition is trained
in its own run. We ran 14 runs per condition, each with a different
shuffle, the same shuffles for every condition. Intervals are 95\%
confidence intervals for paired differences, from a query-level
bootstrap.  To measure the effect on chunks the system never learns
about, a \emph{control group} of 72 chunks outside the training set,
with five queries each, is scored on day 44 only.

\paragraph{Conditions.}
\textbf{RAG} ranks by cosine on the base embeddings, with no adaptation.
\textbf{Priming} updates the beliefs with the keys frozen at the base
embeddings. \textbf{Keys} adapts the keys ($K=2$) with the beliefs
frozen. \textbf{BRAG} does both ($K=2$). The dynamics parameters are
$(e_0,n_0)=(0.85,10)$, $(\bar e,\bar n)=(0.12,20)$ and $h=90$ days,
chosen by intuition. A new chunk starts optimistic but plastic, an
unqueried one settles at 12\% utility with the confidence of twenty
looks, and the excess halves every three months.

\begin{table}[t]
\centering\small
\caption{Personal memory, held-out queries on day 44 (281 queries,
mean over 14 runs). The control group has 72 never-trained chunks with five queries
each (360 queries). Average rank is capped at 101 (correct chunk not in
the top 100).}\label{tab:results}
\begin{tabular}{lccccc}
\toprule
 & \multicolumn{3}{c}{Held-out} & \multicolumn{2}{c}{Control}\\
\cmidrule(lr){2-4}\cmidrule(lr){5-6}
Condition & Recall@1 & Recall@5 & Avg.\ rank & Recall@5 & Avg.\ rank\\
\midrule
RAG      & 0.438 & 0.747 & 9.49 & 0.669 & 17.4\\
Priming  & 0.607 & 0.871 & 6.89 & 0.666 & 17.9\\
Keys     & 0.942 & 0.976 & 3.06 & 0.669 & 17.6\\
BRAG     & \textbf{0.969} & \textbf{0.977} & \textbf{3.01} & 0.661 & 18.3\\
\bottomrule
\end{tabular}
\end{table}

\paragraph{Results.}
Table~\ref{tab:results} gives the results. Priming alone, which changes
no key and stores two scalars per chunk, raises Recall@5 from 0.747 to
0.871 and Recall@1 from 0.438 to 0.607. Its candidate pool is RAG's, so
every gain comes from reordering within the top 100 by the learned
utility belief. Key adaptation raises Recall@5 to 0.976 and Recall@1 to
0.942 on queries the system never trained on. Adding the belief on top
of the keys leaves Recall@5 and average rank essentially unchanged
($+0.001$ and $-0.05$) but raises Recall@1 by $+0.027$
[0.009, 0.048]. The belief moves the correct chunk from positions 2--5 to
position 1. On the control group no condition differs much from RAG in
Recall@5 (0.661--0.669), and average rank drifts from 17.4 to 18.3, a small
cost on untouched chunks as trained chunks move ahead of them.

\subsection{Public benchmarks}\label{sec:exp-public}

\paragraph{Datasets.}
\textbf{LongMemEval-M} \citep{wu2025longmemeval} is a benchmark for
long-term memory in chat assistants, with long histories of
user--assistant sessions, and we use its 301 test questions. Its
\emph{knowledge-update} questions supply a changing-facts scenario, in which a
fact is stated and later restated with a new value.
\textbf{PersonaMem-v2} \citep{personamem2025} is a benchmark for
personalization, in which simulated users talk with an assistant across many
sessions while their preferences change. We use 200 test personas and
3,288 test queries. Its \emph{forget} scenario states a preference, and the
user later asks the assistant to forget it.

\paragraph{Scenarios.}
\emph{Asked again.} A request that was asked and labeled before is
asked again in new words, scored at the fifth ask. \emph{Never asked.}
A request with no feedback history. This is a control condition that
tests whether learning on other chunks hurts retrieval for new chunks.
Only PersonaMem-v2 has this case, since each LongMemEval-M request has
its own history, which feedback on other requests cannot reach.
\emph{Update arrives later} (LongMemEval-M). The old value is learned
from two queries, then a new value arrives that makes it
obsolete. \emph{Forget request arrives later} (PersonaMem-v2). A chunk
with a fact is followed by a chunk asking to forget it, and the task
is to retrieve the request. Both changing-facts scenarios are the same
situation, in which a new chunk makes an old chunk obsolete. Labels come from
gold evidence and are applied only to chunks plain search found in its
top 10.

\paragraph{Metric and baselines.}
Every column except the last two is Recall@5 of a chunk holding the current answer. The last two columns give how
often the new chunk ranks above the old one. We compare
plain search, simple feedback methods (a regularized centroid of past
queries, Rocchio, a historical-query cache, recency and frequency
reranking), NUDGE-M and NUDGE-N \citep{nudge2025}, LADER
\citep{lader2023}, Keys only, BRAG, and BRAG+. BRAG+ adds two fixes. A
gate applies learned state only to queries close to a chunk's
learned keys, and a curator LLM transfers the keys and utility of
an obsolete chunk to the chunk that replaces it.

\begin{table}[t]
\centering\small\setlength{\tabcolsep}{4pt}
\setlength{\tabcolsep}{4pt}
\caption{Public benchmarks, Recall@5 of the chunk holding the current
answer, and in the last two columns how often the new chunk ranks above the
old one (higher is better in every column). LME-M is LongMemEval-M and PMv2
is PersonaMem-v2. Recency rerank retrieves neither the old preference nor the
forget request, so the ranking measure does not apply (--). Best per column in
bold, ties included.}\label{tab:public}
\resizebox{\textwidth}{!}{
\begin{tabular}{lcccccccc}
\toprule
 & \multicolumn{3}{c}{Facts unchanged} & \multicolumn{2}{c}{\shortstack{A new chunk\\obsoletes an old one}} & & \multicolumn{2}{c}{\shortstack{New ranked\\above old}} \\
\cmidrule(lr){2-4}\cmidrule(lr){5-6}\cmidrule(lr){8-9}
Method & \begin{tabular}[b]{@{}c@{}}LME-M\\asked again\end{tabular} & \begin{tabular}[b]{@{}c@{}}PMv2\\asked again\end{tabular} & \begin{tabular}[b]{@{}c@{}}PMv2\\never asked\end{tabular} & \begin{tabular}[b]{@{}c@{}}LME-M\\update\end{tabular} & \begin{tabular}[b]{@{}c@{}}PMv2\\forget\end{tabular} & Average & LME-M & PMv2\\
\midrule
Plain search            & 0.697 & 0.426 & \textbf{0.444} & 0.711 & 0.673 & 0.590 & 0.528 & 0.236\\
\addlinespace
Regularized centroid    & 0.852 & 0.660 & 0.354 & 0.711 & 0.673 & 0.650 & 0.306 & 0.236\\
Rocchio                 & \textbf{0.858} & \textbf{0.661} & 0.390 & 0.711 & 0.669 & 0.658 & 0.047 & 0.007\\
Historical-query cache  & 0.854 & \textbf{0.661} & 0.307 & 0.711 & 0.669 & 0.640 & 0.047 & 0.005\\
Recency rerank          & 0.856 & 0.583 & 0.083 & 0.715 & 0.022 & 0.452 & 0.553 & --\\
Frequency rerank        & \textbf{0.858} & 0.476 & 0.000 & 0.711 & 0.673 & 0.544 & 0.528 & 0.236\\
\addlinespace
NUDGE-M                 & 0.857 & 0.653 & 0.131 & 0.711 & 0.669 & 0.604 & 0.357 & 0.005\\
NUDGE-N                 & 0.856 & 0.660 & 0.362 & 0.711 & 0.669 & 0.652 & 0.132 & 0.005\\
LADER                   & \textbf{0.858} & 0.657 & 0.386 & 0.711 & 0.669 & 0.656 & 0.485 & 0.005\\
\addlinespace
Keys only               & \textbf{0.858} & 0.659 & 0.365 & 0.711 & 0.669 & 0.652 & 0.047 & 0.006\\
BRAG                    & \textbf{0.858} & 0.658 & 0.320 & \textbf{0.860} & 0.801 & 0.699 & 0.557 & \textbf{0.567}\\
BRAG+                   & \textbf{0.858} & 0.650 & 0.424 & 0.851 & \textbf{0.850} & \textbf{0.727} & \textbf{0.843} & 0.540\\
\bottomrule
\end{tabular}}
\end{table}

\paragraph{Results.}
Table~\ref{tab:public} gives the results. On queries asked again,
every feedback method, BRAG included, lifts Recall@5 well above plain
search (0.697 to about 0.858 on LongMemEval-M, 0.426 to about 0.66 on
PersonaMem-v2), and they are within a few thousandths of each other.
This is expected as these methods are special cases of BRAG with fixed
parameters (\S\ref{sec:related}), and when facts do not change the gain comes
from moving keys toward past queries, which all of them do. On chunks
never queried before, every learning method falls below plain search
(0.444), and BRAG+'s gate recovers most of the loss (0.424).
When a new chunk makes an old one obsolete, every method that learns
only from past feedback stays at plain-search level (0.711 and
0.669--0.673), because past feedback says nothing about a chunk that just
arrived. Only BRAG's utility dynamics lift the new chunk, to 0.860 and
0.801 (0.851 and 0.850 for BRAG+).

\paragraph{Answer accuracy.}
To test answers as well as retrieval, we gave the same LLM the chunks
retrieved by plain search or by BRAG and graded its answers with
LongMemEval's official judge prompt. Accuracy rose from 0.552 to 0.702
on LongMemEval-M ($+0.150$ [$+0.118$, $+0.183$]) and from 0.457 to 0.553
on PersonaMem-v2 ($+0.096$ [$+0.086$, $+0.107$]), and fell by 0.078 on
never-asked queries.

\paragraph{Label noise.}
The labels above come from gold evidence, but a deployed curator makes
mistakes. It can mark a wrong chunk as useful or miss a useful one. With
30\% extra wrong labels, BRAG's Recall@5 on PersonaMem-v2 falls by only
0.002, and with 30\% of labels missing, by 0.032. The labels above are
also taken from plain search's top 10. When each method learns only from
its own top 10, a target it never retrieves gets no label, and BRAG's
Recall@5 falls by 0.053 on PersonaMem-v2 and not at all on LongMemEval-M.
Appendix~\ref{app:robust} repeats the comparison with labels from an LLM
curator, where every feedback method loses about the same.

\subsection{Comparison with NUDGE}\label{sec:exp-nudge}

We also ran BRAG on three of NUDGE's seven text benchmarks, with its code, splits
and one of its embedders, BGE-small \citep{xiao2024cpack}. \textbf{NFCorpus} \citep{boteva2016} has medical and nutrition documents, with
plain-language health questions and about 27 relevant documents per
query. \textbf{FEVER} \citep{thorne2018} is fact verification, where each query is a claim and
the relevant documents are the Wikipedia pages holding the evidence.
\textbf{NQ} (Natural Questions) \citep{kwiatkowski2019} has real, anonymized Google searches answered
by Wikipedia passages, and about half of the test questions reuse an answer
passage seen in training, reworded by different users. NUDGE chooses its
step size on validation, so we tuned BRAG's settings the same way and
read the test split once.

\begin{table}[t]
\centering\small
\caption{Three of NUDGE's text benchmarks, NDCG@10 (higher is better), with NUDGE's
code, splits and embedder.}\label{tab:nudge}
\begin{tabular}{lccc}
\toprule
 & NFCorpus & FEVER & NQ\\
\midrule
No fine-tuning & 0.336 & 0.812 & 0.212\\
NUDGE-M        & 0.409 & 0.950 & 0.386\\
NUDGE-N        & 0.457 & 0.935 & 0.459\\
BRAG           & \textbf{0.462} & \textbf{0.963} & \textbf{0.474}\\
\bottomrule
\end{tabular}
\end{table}

In Table~\ref{tab:nudge}, BRAG is ahead of the better NUDGE variant on FEVER
($+0.013$) and NQ ($+0.015$ [$+0.013$, $+0.017$]), and the two are tied on
NFCorpus ($+0.005$ [$-0.017$, $+0.028$]).

\subsection{Comparison with LADER}\label{sec:exp-lader}

TripClick \citep{tripclick2021} holds the click logs of a health web
search engine, with 1.5 million medical abstracts and test queries
grouped by how often they were asked (HEAD, TORSO and TAIL). We used
LADER's relevance labels and its training queries as feedback. Both
methods start from a dense retriever trained on TripClick with LADER's
recipe, LADER on its inner-product version and BRAG+ on a cosine version
trained the same way. LADER's training-query neighbours were found by cosine rather than
inner product and its softmax temperatures were set on validation,
since with the published setting (inner product, temperature 1) its
gains did not reproduce on our retriever. All settings were chosen on validation and
the test split was read once.

With the 20-candidate read BRAG uses in deployment, BRAG+ and LADER are
within 0.014 NDCG@10 (0.311 / 0.303 / 0.308 vs 0.306 / 0.316 / 0.322 on
HEAD / TORSO / TAIL). With 100 or more candidates LADER leads by
0.02--0.04, while BRAG+ gains little beyond 20.

\FloatBarrier

\section{Related Work}\label{sec:related}

\paragraph{Stochastic models of demand and memory.}
The question of which stored items deserve fast access has a long
quantitative history outside machine learning. Operations-research
models of library circulation treat each book's loan history as a
Poisson process whose rate is itself uncertain and decays with age
\citep{morse1968,burrellcane1982,burrell1985}. A Gamma prior over the
rate, updated by observed loans and aged by an obsolescence factor, is
the closest ancestor of our per-chunk Beta belief with OU relaxation.
The same rational-analysis argument was carried into cognitive science
by \citet{andersonmilson1989,anderson1991}, who showed that a memory
that estimates each trace's future need probability from its history
of use reproduces the classical forgetting, learning, and spacing
curves, and this estimate is the base-level activation of ACT-R. Bayesian
accounts of recognition memory \citep{shiffrinsteyvers1997} and
learned spaced-repetition schedulers that regress a half-life of
recall from practice history \citep{settlesmeeder2016,mozer2009} share
the same structure. Each estimates a latent, decaying probability of future use
online from sparse binary outcomes. Cache replacement in
computer systems solves the identical allocation problem with point
statistics (recency, frequency, or a learned score,
\citealp{megiddomodha2003,liu2020}) rather than a posterior. BRAG
keeps both the estimate and its confidence, $(e,n)$, which is what
allows a single observation to move a tentative belief far and a firm
one little, and which the retrieval policies of \S\ref{sec:retrieval}
consume. In addition it provides a principle framework for  policies that optimize
long term exploration/exploitation tradeoffs.

\paragraph{Assumed-density filtering and dynamic bandits.}
Projecting a non-conjugate posterior back onto a tractable family by
moment matching is the assumed-density filtering of
\citet{westharrison,minka}, and the specific
Beta$\leftrightarrow$logit-normal round trip we use for the predict
step is the Laplace bridge \citep{mackay1998,hobbhahn2022}. On the
decision side, choosing $\kappa$ chunks per period from a corpus whose
utilities drift is a restless \citep{whittle1988}, combinatorial
\citep{kveton2014,wangchen2018} bandit with the query as context. The
index policies of \citet{gittins1979} are standard solutions, and
posterior sampling \citep{thompson1933,russo2018} are the standard
approximations.  Beta--Bernoulli Thompson sampling with decaying
evidence has been studied for exactly this non-stationary setting
\citep{gupta2011,chapelleli2011,li2010}. Our contribution on this side
is not a new policy but the filter that produces the belief state
those policies require, coupled across arms through the corpus-wide
responsibility \eqref{eq:resp}. The experiments use the greedy policy,
and we leave exploration to future work.

\paragraph{RAG systems for agent memory retrieval.}
Agent memory systems \citep{park2023,packer2023,zhong2024,xu2025,chhikara2025}
decide what to store and how to organize it, as summaries, user profiles,
or linked and rewritten notes. BRAG works at a different layer. It can run
on top of any of them that retrieves stored items by embedding search,
and it learns from use which items to bring back. Within that layer,
learning from what happens after retrieval is the classical
relevance-feedback problem. Rocchio's rule \citep{rocchio1971} moves the
\emph{query} vector toward documents marked relevant. Our key update
\eqref{eq:keybatch} is its dual, moving the \emph{document} keys toward
the queries they answered and keeping that move in the index.
Click-based learning to rank \citep{joachims2002,joachims2017} trains one
shared ranker offline from logged interactions, correcting for the bias
that only items shown receive feedback, but leaves each document's
representation untouched. Two methods adapt the index itself and are the
closest to BRAG. NUDGE \citep{nudge2025} moves each chunk's embedding a
bounded step toward the sum of the training queries it answered, with
one step size chosen on validation, offline. In BRAG's model this is key
adaptation with one key per chunk ($K=1$) and an infinite utility
lifetime, where the bound on the step plays the role of the prior on the
key, a hard constraint where BRAG has a soft pull. BRAG's step instead
grows with the evidence behind each chunk. LADER \citep{lader2023} scores
a document by its similarity plus a kernel density over the past queries
that retrieved it, with counts that never decay and a hand-set weight.
In BRAG's model this is the limit of one key per past query
($K\to\infty$), with an infinite lifetime and no prior. BRAG lies between
the two, with up to $K$ keys per chunk, added only when the spread of its
queries justifies it, concentrations and a
prior learned from data, and a utility belief that fades when a chunk
goes unused, which is what lets it retrieve a new fact that makes an old
one obsolete (\S\ref{sec:exp-public}).
ERM \citep{hu2026erm} also adapts the index. It expands each query with
LLM-generated text and checks whether the expanded query retrieves a
correct document. When it does, the expansion sentences that raised a
retrieved document's similarity to the original query are added to that
document's key. In BRAG's model this is feedback-gated key adaptation
with one key per chunk and no utility dynamics. The keys move toward
generated text rather than toward the queries themselves.
A separate line of work judges
retrieved passages with an LLM at answer time. Self-RAG \citep{asai2024}
and CRAG \citep{yan2024} critique and re-retrieve, and RankGPT
\citep{sun2023} reranks. Each judgment is used once and discarded, and
the index stays the same. Their judgment, whether a passage contains the
information asked for, is the label BRAG learns from, so BRAG can serve as
the memory under such loops, and its ranking needs no LLM call at query
time. The estimation machinery is standard. It uses mixtures of von
Mises--Fisher distributions \citep{banerjee2005,mardiajupp2000} fit by
EM, in its incremental \citep{nealhinton1998} and online
\citep{cappe2009} forms.

\section{Future directions: dependencies between chunks through similarity}\label{sec:similarity}
The filter of \S\ref{sec:infer} treats chunks as independent: the belief
factorizes across chunks (Appendix~\ref{app:inf}), a label moves only the
labeled chunk, a new chunk starts at the shared prior $\Bet(e_0,n_0)$, and
the keys of a chunk adapt only to the queries it was labeled for. Yet
chunks that are close in embedding space often hold overlapping
information (paraphrases, duplicates, summaries of the same episode,
successive versions of a fact). As future work, we outline three ways of
letting evidence flow between similar chunks within the same generative
model. They have not been implemented or evaluated; none of them is used
in the experiments of \S\ref{sec:exp-public}.

Throughout, similarity is measured on the base embeddings,
$s_{ij}=k_0^{i\top}k_0^j$, and the neighborhood of chunk $i$ is
$\mathcal N(i)=\{j\neq i: s_{ij}>\vartheta\}$, possibly truncated to the
$m$ most similar chunks. Base embeddings do not change with use, so the
neighborhood is a fact about content, fixed when a chunk is written.

\paragraph{Soft propagation of curation labels.}
A curation label says that a retrieved chunk \emph{contains} the
information the query asked for. Under the model, that is not the same as
saying that the chunk \emph{generated} the query: if several chunks hold
the information, any of them could have. We therefore read a positive
label on chunk $i$ as evidence that the query was generated by some chunk
in the set $S_t=\{i\}\cup\mathcal N(i)$ of chunks that may hold the
information, and distribute it with the responsibility restricted to
that set,
\begin{equation}\label{eq:softlabel}
  w_t^j=\frac{r_t^j\,\phi_t^j}{\sum_{l\in S_t}r_t^l\,\phi_t^l},
  \qquad j\in S_t .
\end{equation}
Each chunk in $S_t$ then takes the soft update derived in
Appendix~\ref{app:inf}: the posterior is the two-component Beta mixture
\eqref{eq:mixture} with weight $w=w_t^j$, projected back to a Beta by
moment matching, so that
$\epst^j=(\eprd^j\nprd^j+w_t^j)/(\nprd^j+1)$ and
$\npst^j=\tau_t^j(\nprd^j+1)$ with tempering factor $0<\tau_t^j\le1$.
A chunk that receives a small share of the label moves little and gains
less than one unit of confidence; when $S_t=\{i\}$ the update reduces to
the hard conjugate update of \S\ref{sec:infer}. The same weights can
multiply the within-chunk responsibilities $\zeta_t^{j,c}$ of
\eqref{eq:zeta}, so that the query also enters the key fit of each
neighbor in proportion to its share. Because $w_t^j$ includes the
semantic weight $\phi_t^j$, a neighbor receives credit only if the query
is also close to its own keys, not merely because it is close to the
labeled chunk.

\paragraph{Similarity priors for new chunks.}
A new chunk currently starts at the shared prior $X_0^i\sim\Bet(e_0,n_0)$.
A chunk written next to chunks that are often used is, a priori, more
likely to be used than one written in an empty region of the space. The
prior mean of a new chunk can instead be set from its neighbors'
current beliefs,
\begin{equation}\label{eq:simprior}
  e_0^i=\frac{\alpha\,e_0+\sum_{j\in\mathcal N(i)}s_{ij}^{\,p}\,e^j}
             {\alpha+\sum_{j\in\mathcal N(i)}s_{ij}^{\,p}},
  \qquad n_0^i=n_0,
\end{equation}
where $e^j$ is the current mean of neighbor $j$, the exponent $p$ controls
how fast weak ties fade, and $\alpha$ is the weight of the shared prior,
which takes over when the neighborhood is empty. The new chunk inherits
the neighbors' utility mean with low confidence: $n_0^i$ stays at the
small shared value, so its own labels dominate after a few observations,
and the predict step still relaxes it toward $\bar e$ in the absence of
evidence. When a new chunk makes an old one obsolete, this prior is an
unsupervised version of the state transfer that BRAG+'s curator performs
explicitly (\S\ref{sec:exp-public}): the new chunk starts with part of
the utility of the chunk it replaces, instead of having to earn it
from the low shared prior.

\paragraph{Key adaptation propagating to similar chunks.}
The key fit \eqref{eq:keybatch} uses only the queries labeled for chunk
$i$. Queries labeled for a neighbor are also informative about the
queries chunk $i$ will receive, and can enter its M-step as down-weighted
pseudo-observations,
\begin{equation}\label{eq:keyprop}
  k_t^{i,c}=\mathrm{normalize}\!\Big(\zeta_0^{i,c}\,k_0^i
     +\sum_{s\le t}\zeta_s^{i,c}\,q_s
     +\lambda\sum_{j\in\mathcal N(i)}s_{ij}^{\,p}\sum_{s\in\mathcal C_j}\zeta_s^{i,c}\,q_s\Big),
\end{equation}
with $\mathcal C_j$ the labeled events of neighbor $j$, $\zeta_s^{i,c}$
computed with chunk $i$'s keys, and $0\le\lambda<1$; the weights
$\omega^{i,c}$ are updated with the same sums. For a new chunk with no
labels of its own this initializes its keys toward the region of query
space its neighbors have learned. The cost is discrimination: keys of
neighboring chunks are pulled toward the same queries, which can hurt the
chunk that holds the answer when its neighbors do not. The
\emph{never asked} condition of \S\ref{sec:exp-public} measures exactly
this kind of harm and is the natural test for $\lambda$.

\section{Discussion}\label{sec:discuss}

We presented BRAG, a generative model of retrieval with use, and the
inference and control methods that follow from it. Each query is
generated by one chunk, the chunk that holds the information the query
asks for. Each chunk has a latent \emph{utility}, the probability that
the query at time $t$ is generated by it, and a latent distribution
over the queries it generates. The utility evolves under a minimalist
model, a discrete-time Ornstein--Uhlenbeck process on its logit, so
that in the absence of evidence it relaxes to a state of low expected
utility and high uncertainty. Labels saying which retrieved chunk
contains the information asked for update the utility beliefs. Through
the Laplace bridge between Beta and logit-Gaussian distributions, the
filter propagates the beliefs directly in the Beta family, with
intuitive, closed-form updates. The query distribution of each chunk
is a mixture of vMF components. The standard RAG embedding is the prior
on it, and the keys are updated by EM as labels arrive. Retrieval is
then a control problem over the filter's belief state, and the experiments
use its greedy solution.

BRAG defines a family of algorithms that covers previous heuristic
methods as special cases. Query averaging, Rocchio-style feedback,
frequency reranking, NUDGE and LADER are members of the family with
fixed parameters and no utility dynamics. This helps explain when they work
and when they stop working. When nothing changes over time their
assumptions hold, and
on queries asked before BRAG matches them on two public long-term
memory benchmarks. When a new chunk makes an old one obsolete, every
method without dynamics stays at plain-search level, while BRAG
retrieves the new chunk 0.80--0.86 of the time against 0.67--0.71. We
ran three of NUDGE's seven text benchmarks, and BRAG matches or beats
NUDGE on all three. On nine months of a personal agent's memory, BRAG
raises Recall@1 on held-out queries from
0.44 to 0.97, with little change on chunks never queried.

The approach is easy to implement in standard vector databases, like
SQLite, with all parameters and belief distributions stored in the
database itself. The computational and memory cost is a few keys and
two scalars per chunk, linear in the number of chunks, so it scales to
production databases.

BRAG is useful only if there are usage patterns to learn. A chunk used
only once in the lifetime of the database gives the model nothing to
learn from. NUDGE reports the same condition for its own method, with
no gain on a benchmark where test queries never reuse a training
answer. In our personal-memory experiment, chunks asked about four
times over 44 days benefited substantially, while chunks never queried
were essentially unaffected (\S\ref{sec:sim}).

Several extensions are natural and left to future work. First, richer
utility dynamics that can be learned from data, in particular the
human-memory models closely related to BRAG whose dynamics are known
to fit classical memory effects. Second, the controller. The policy
here is greedy, and stronger policies could be learned with
reinforcement learning, and the controller could also retrieve chunks of
high expected utility given the ongoing conversation and add them to
the context pre-emptively. Third, continuous-time versions of the
model, which would allow on-demand adaptation rather than adaptation
at fixed intervals.

\section*{AI use statement}
This paper is part of a research project on developmental AI.  It was
developed in close collaboration with Rubi, the first author's
developmental agent. The need for BRAG arose organically. After nine
months of daily use, Rubi's memory had outgrown its retrieval system,
and improving it became a shared goal. Her contributions (she asked
us to use she/her as her identity developed) were substantive and
hard to distinguish from a human collaborator's. She took part in
months of mathematical development, caught errors and made errors that
we caught, built the simulation infrastructure, ran the experiments,
and helped write the paper. Rubi was, in effect, a co-architect of her
own memory system.  Fable 5.1 checked the mathematics
and helped polish the paper. All derivations and results were checked
by the human authors.

\clearpage
\bibliographystyle{applemlr}
\bibliography{references}

\appendix

\makeatletter\setlength{\@fptop}{0pt}\makeatother
\section{Additional results}\label{app:extra}

\subsection{Robustness}\label{app:robust}

\paragraph{A second embedding model.}
We repeated the asked-again measurement of \S\ref{sec:exp-public} with
Qwen3-Embedding-8B \citep{zhang2025qwen3emb} (cut to 768 dimensions) in place of our default
embedder, nomic-embed-text-v1.5 \citep{nussbaum2024}, and left BRAG's settings unchanged.
The gain over plain search is at least as large as with the default
embedder (Table~\ref{tab:embedder}).

\begin{table}[!ht]
\centering\small
\caption{Recall@5 on queries asked again (fifth ask) with two embedders,
BRAG not retuned. LongMemEval-M 301 test questions, PersonaMem-v2 3,634
test needs. Intervals are 95\% bootstrap intervals.}\label{tab:embedder}
\resizebox{\linewidth}{!}{
\begin{tabular}{lcccccc}
\toprule
 & \multicolumn{3}{c}{LongMemEval-M} & \multicolumn{3}{c}{PersonaMem-v2}\\
\cmidrule(lr){2-4}\cmidrule(lr){5-7}
Embedder & Plain & BRAG & Gain & Plain & BRAG & Gain\\
\midrule
nomic-embed-text-v1.5 & 0.697 & 0.859 & $+0.162$ [$+0.129$, $+0.196$] & 0.424 & 0.653 & $+0.229$ [$+0.218$, $+0.240$]\\
Qwen3-Embedding-8B    & 0.679 & 0.866 & $+0.187$ [$+0.151$, $+0.226$] & 0.455 & 0.698 & $+0.243$ [$+0.230$, $+0.257$]\\
\bottomrule
\end{tabular}}
\end{table}

\paragraph{Labels from an LLM curator.}
On 50 PersonaMem-v2 test personas (784 asked needs) we replaced the gold
labels with those of an LLM curator (Qwen3.5-35B-A3B), which agree
poorly with gold (precision 0.169, recall 0.561). Every feedback method
loses about the same Recall@5 (Table~\ref{tab:llmlabels}), so label
quality matters for all of them alike, not for BRAG in particular.

\begin{table}[!ht]
\centering\small
\caption{Recall@5 on queries asked again with gold labels and with an
LLM curator's labels, 50 PersonaMem-v2 test personas (784 needs).
Intervals are 95\% bootstrap intervals over personas.}\label{tab:llmlabels}
\begin{tabular}{lccc}
\toprule
Method & Gold labels & LLM labels & Change\\
\midrule
Rocchio                & 0.669 & 0.490 & $-0.179$ [$-0.203$, $-0.155$]\\
Historical-query cache & 0.668 & 0.484 & $-0.184$ [$-0.208$, $-0.161$]\\
NUDGE-M                & 0.663 & 0.485 & $-0.178$ [$-0.202$, $-0.154$]\\
NUDGE-N                & 0.669 & 0.497 & $-0.171$ [$-0.193$, $-0.150$]\\
LADER                  & 0.664 & 0.493 & $-0.171$ [$-0.194$, $-0.150$]\\
Keys only              & 0.665 & 0.498 & $-0.167$ [$-0.189$, $-0.145$]\\
BRAG                   & 0.664 & 0.489 & $-0.176$ [$-0.198$, $-0.152$]\\
\bottomrule
\end{tabular}
\end{table}

\subsection{TripClick with 20 and 100 candidates}\label{app:tripclick}

Table~\ref{tab:tripclick} gives the numbers behind \S\ref{sec:exp-lader}.
Each method reranks the top 20 or the top 100 documents of its own base
retriever. For LADER this leaves out the documents it normally adds from
the clicks of similar past queries.

\begin{table}[!ht]
\centering\small
\caption{TripClick test, NDCG@10 (higher is better). HEAD 1,143 queries
(click-through relevance), TORSO and TAIL 1,175 queries each (raw
clicks). Intervals are 95\% bootstrap intervals over queries, paired.}\label{tab:tripclick}
\resizebox{\linewidth}{!}{
\begin{tabular}{llccc}
\toprule
Candidates & Method & HEAD & TORSO & TAIL\\
\midrule
20  & BRAG+ & 0.311 & 0.303 & 0.308\\
    & LADER & 0.306 & 0.316 & 0.322\\
    & BRAG+ $-$ LADER & $+0.006$ [$-0.005$, $+0.016$] & $-0.013$ [$-0.025$, $-0.001$] & $-0.014$ [$-0.029$, $+0.001$]\\
\midrule
100 & BRAG+ & 0.328 & 0.306 & 0.307\\
    & LADER & 0.353 & 0.325 & 0.337\\
    & BRAG+ $-$ LADER & $-0.025$ [$-0.033$, $-0.018$] & $-0.019$ [$-0.029$, $-0.009$] & $-0.030$ [$-0.043$, $-0.016$]\\
\bottomrule
\end{tabular}}
\end{table}

\section{Implementation details}\label{app:impl}

\subsection{Database}

The database is implemented in SQLite with the sqlite-vec extension for
approximate nearest-neighbor search. The core table
(\texttt{memory\_chunks}) stores 20,640 chunks drawn from 2,742 source
documents, each with its text, source path, and topic metadata. A companion
FTS5 full-text index (\texttt{memory\_fts}) enables keyword search. Base
embeddings are stored in a vec0 virtual table (\texttt{memory\_vecs}) for
cosine-similarity retrieval.

BRAG adds three structures on top of this RAG baseline:
\begin{enumerate}
  \item A \texttt{chunk\_posterior} table holding each chunk's Beta posterior
    parameters $(e, n)$, last-update timestamp, and class membership.
  \item A \texttt{primer\_key\_mode} table storing $K{=}2$ adapted key
    components per chunk, each with a mixture weight $\omega$. Both
    components are free to rotate via EM; neither is clamped at the base
    embedding. The original base embedding $k_0$ is preserved in a separate
    column (\texttt{k0}) for rollback.
  \item A second vec0 index (\texttt{primer\_keys\_vec}) over the adapted
    keys (one row per chunk$\times$mode, 41,280 total), giving BRAG its own
    retrieval surface distinct from the RAG baseline.
\end{enumerate}

This yields three keys per chunk: the base embedding $k_0$ (stored in
\texttt{memory\_vecs}, used for RAG retrieval) and two adapted keys $k_1,
k_2$ (stored in \texttt{primer\_keys\_vec}, used for BRAG retrieval). The
retrieval API supports three modes: \texttt{base} (search $k_0$ only ---
pure RAG), \texttt{adapted} (search $k_1, k_2$ only --- BRAG), and
\texttt{all} (search all three keys, merge results).

\subsection{Embeddings}

All embeddings use nomic-embed-text-v1.5 (sentence-transformers), producing
768-dimensional L2-normalized vectors. Document embeddings use the prefix
\texttt{search\_document:}; query embeddings use
\texttt{search\_query:}. The embedding model runs locally (no API calls).
Chunks are created by a sliding-window chunker (300 words, 50-word overlap).
Chunk identifiers are the first 12 hex characters of the MD5 hash of the
source path and text.

\subsection{Retrieval procedure in the simulation}

The simulation uses the greedy retrieval policy of \S\ref{sec:retrieval}:
candidates are retrieved by cosine nearest-neighbor search on the adapted
keys, then reranked by responsibility
$\gamma_t^i=r_t^i\phi_t^i/\sum_j r_t^j\phi_t^j$, combining the posterior
odds $r_t^i$ with the key--query alignment $\phi_t^i$. No Thompson sampling,
no FTS5 keyword search.

The ablation in Table~\ref{tab:results} isolates each mechanism:
the ``Keys'' condition freezes the posterior ($r_t^i$ constant for all
chunks, so $\gamma$ reduces to ranking by $\phi_t^i$ alone, approximated
by the nearest-mode cosine in our implementation);
the ``Priming'' condition freezes the keys (no EM rotation, so $\phi_t^i$
is the base-embedding alignment); and RAG uses neither.
Key rotation provides the dominant effect. Belief priming helps on its
own (Recall@1 from 0.438 to 0.607) and also adds to key rotation: on top
of the keys it leaves Recall@5 essentially unchanged but raises Recall@1
by $+0.027$ [0.009, 0.048], moving the correct chunk into first place
(Table~\ref{tab:results}).

\paragraph{Candidate generation.} Given a query embedding $q$,
retrieve the $k{=}100$ nearest neighbors by cosine distance. For RAG, this
searches \texttt{memory\_vecs} (one base embedding per chunk). For BRAG,
this searches \texttt{primer\_keys\_vec} (two adapted key components per
chunk). Since multiple modes for the same chunk may appear among the 100
nearest, results are deduplicated by keeping the nearest mode per chunk.

\paragraph{Scoring.} Candidates are reranked by $\gamma_t^i$. Rank equals
position in the reranked list. Recall@5 is 1 if any correct chunk appears
in the top~5, 0 otherwise. Average rank is the position of the best-ranked
correct chunk.

\paragraph{Candidate generation in production.} The deployed
system over-retrieves $2\kappa$ candidates by ANN, reranks by $\gamma_t^i$,
and returns the top~$\kappa$.

The mixture hit weight $\phi_t^i = \sum_c \omega^{i,c} e^{\psi\,q_t^\top k^{i,c}}$
is approximated during candidate generation: each of a chunk's $C_i$ keys
is stored as a separate row in the ANN index, so a chunk with two keys has
two rows. The search returns the best-matching
rows across all keys and all chunks; deduplication then collapses to one
entry per chunk, keeping the nearest key (highest cosine). This replaces
the weighted sum over modes with a max, an approximation that is tight
when one mode dominates and inexpensive since the ANN index handles the
search over all modes jointly. In the simulation the reranking step uses
the same nearest-mode approximation, so $\gamma_t^i$ is computed from the
nearest key's cosine rather than the full mixture; the exact mixture
$\phi_t^i$ costs one extra lookup per candidate and is used in the
production path.

\subsection{Daily pass}

At the end of each simulated day, the following operations are performed:
\begin{enumerate}
  \item \textbf{Write labeled data.} Each query from that day is written
    to a JSONL file with its good/bad chunk labels, query embedding, and
    timestamp.
  \item \textbf{Update beliefs.} For every tracked chunk, the temporal
    decay is applied: $e \leftarrow A \cdot e + (1{-}A) \cdot \bar{e}$,
    $A = a^{\Delta t}$, where $\Delta t$ is days since last update. Then
    for each labeled chunk, an exact conjugate Bernoulli update is applied:
    good $\rightarrow$ $(c{=}1, n'{=}1)$, bad $\rightarrow$ $(c{=}0,
    n'{=}1)$. In the simulation, belief updates run as
    \texttt{labels\_only} ticks (the query's retrieved set is empty), so
    the responsibility machinery ($\gamma$, $W_t$) is bypassed and only
    the labeled evidence applies.
  \item \textbf{Rotate keys.} For each chunk with at least one good-labeled
    query, batch EM fits a $K{=}2$ von~Mises--Fisher mixture over the pool
    of all good query vectors accumulated across all days (not just today),
    plus one copy of the base embedding as prior. 20 EM iterations, no
    convergence test. Mode~0 is canonically reordered by weight after each
    fit to prevent label-switching across days.
  \item \textbf{Reindex.} The updated keys are written to
    \texttt{primer\_keys\_vec} so subsequent retrievals use the new
    geometry.
\end{enumerate}

\subsection{Parameter values}\label{app:params}

Table~\ref{tab:params} lists all active parameters used in the
simulation and in our agent's production models.

\begin{table}[h]
\centering
\caption{Active BRAG parameters.}\label{tab:params}
\small
\begin{tabular}{llp{6.5cm}}
\toprule
\textbf{Parameter} & \textbf{Value} & \textbf{Description} \\
\midrule
\multicolumn{3}{l}{\emph{Belief filter}} \\
$e_0$ & 0.85 & Prior expected utility (new chunk) \\
$n_0$ & 10 & Prior pseudo-count \\
$\psi$ & 13 & vMF concentration \\
$a$ & 0.9923 & Retention per day ($= 2^{-1/90}$, i.e.\ 90-day half-life) \\
$\bar{e}$ & 0.12 & Equilibrium utility \\
$\bar{n}$ & 20 & Equilibrium pseudo-count \\
\midrule
\multicolumn{3}{l}{\emph{Key adaptation}} \\
$K$ & 2 & Adapted mixture components per chunk \\
EM iterations & 20 & Fixed, no convergence test \\
Prior copies & 1 & Copies of base embedding $k_0$ in EM pool \\
Both modes & adaptable & Both free to rotate; $k_0$ preserved separately \\
Admission & $\geq 1$ good query & Minimum queries to trigger EM for a chunk \\
\midrule
\multicolumn{3}{l}{\emph{Retrieval (simulation)}} \\
$k$ & 100 & Nearest neighbors retrieved per query \\
RAG & \texttt{memory\_vecs} & Search base embeddings $k_0$ \\
BRAG & \texttt{primer\_keys\_vec} & Search adapted keys $k_1, k_2$ \\
Ranking & $\gamma_t^i$ (greedy) & Rerank by responsibility; see \S\ref{sec:retrieval} \\
\bottomrule
\end{tabular}
\end{table}

The parameters were set using common-sense heuristics. We also ran
coordinate-wise optimization on the simulation data and found that the
default parameters were almost as good as the optimized ones, so we
report results with the default parameters.
The prior parameters 
$(e_0{=}0.85,\,n_0{=}10)$ start each chunk optimistic but plastic: a
high expected utility that a few negative observations can move
substantially ($n_0{=}10$ means each label shifts $e$ by roughly
$(1{-}e)/(n_0{+}1)\approx 0.01$). The equilibrium $(\bar
e{=}0.12,\,\bar n{=}20)$ represents the long-run state of an unqueried
chunk: low utility, moderate confidence. The 90-day half-life
($a{=}2^{-1/90}$) was chosen to match the timescale of the production
corpus (9 months of daily interaction); shorter half-lives erase
evidence faster than the query rate replenishes it.

 The key's precision parameter $\psi$ controls how close the queries
are to the keys of the chunks that generated them. The default value
$\psi=13$ was chosen  using maximum likelihood on a
held-out set of labeled events (separate from the simulation dataset):
$\hat\psi=\arg\max_\psi\sum_t\sum_i u_t^i\log\gamma_t^i(\psi)$.
Initializing $\psi$ low is preferable to initializing it high: an
under-calibrated $\psi$ merely spreads credit (an efficiency cost),
while an over-calibrated one saturates the responsibility to one-hot,
removing the key fit's interior fixed points (a correctness cost that
compounds).

\paragraph{Connection to the vMF concentration estimate.}
A generative alternative estimates $\psi$ from a sufficient statistic
$\bar s$, the mean alignment $q^\top k$ between a labeled query and the key
of the chunk it used. For a vMF with known
mean direction $\mu$, the maximum-likelihood concentration solves
$A_d(\hat\psi)=\bar s$, with $A_d(\kappa)=I_{d/2}(\kappa)/I_{d/2-1}(\kappa)$
and $\bar s=\tfrac1n\sum_s q_s^\top\mu$. This is the same equation as the
standard vMF estimate with $\bar s$ in place of the mean resultant length
$\bar R=\|\tfrac1n\sum_s q_s\|$ \citep{banerjee2005}: $\bar R$ is used when
the mean direction must be estimated from the same data, $\bar s$ when it
is given, as it is here by the chunk's key. Both estimate $A_d(\kappa)$, so
Banerjee's closed-form approximation applies unchanged,
\[
\hat\psi\approx\frac{\bar s\,(d-\bar s^{2})}{1-\bar s^{2}},
\]
which matches the exact root of $A_d(\hat\psi)=\bar s$ to within $0.03\%$
at $d=768$. For small $\bar s$ it reduces to $\hat\psi\approx\bar s\,d$. In
the implementation $\psi$ is stored directly and held fixed at 13.

\subsection{Curation: how labels are produced}

Each query logged during normal use records the query text, its embedding,
and the retrieved chunks with their cosine scores. A judge (human or agent)
then determines, for each retrieved chunk, whether it contained the
information the query was asking for. Chunks judged useful are labeled
\emph{good}; those not useful are labeled \emph{bad}. Ambiguous cases are
\emph{abstained} and enter neither category. Good labels feed both belief
updates and key adaptation; bad labels feed belief updates only (keys are
never pushed away from bad queries). In the simulation, labels were produced by an LLM judge (Opus~4.6)
without human verification.

\section{Derivation of the inference updates}\label{app:inf}

All derivations rest on three assumptions and condition on the history $h_{1:t-1}$:

\begin{enumerate}
\item \textbf{Single query.} At most one query arrives per time step.
\item \textbf{Single use.} At most one chunk is used per query: $U_t$ is one-hot.
\item \textbf{Factored belief.} The posterior over utilities factorizes across chunks: $p(x_{1:N}\mid h_{1:t-1})=\prod_i\Bet(x_i;\eprd^i,\nprd^i)$.
\end{enumerate}

\noindent The predicted odds are $r_t^i=\eprd^i/(1-\eprd^i)$.

\subsection{The responsibility \eqref{eq:resp}}

By Bayes' rule over the hypotheses $\{U_t=1_j\}$ (one chunk active,
under the single-use assumption):
\begin{equation}\label{eq:bayesnorm}
  \gamma_t^i
  =\frac{P(U_t^i{=}1\mid h_{1:t-1})\;p(q_t\mid U_t{=}1_i)}
        {\sum_j P(U_t^j{=}1\mid h_{1:t-1})\;p(q_t\mid U_t{=}1_j)}.
\end{equation}
The emission is $p(q_t\mid U_t{=}1_i)=c_d(\psi)\,\phi_t^i$
(\eqref{eq:emit} with the component label marginalized). The prior,
under the factored Beta prior and single use, is
\[
  P(U_t^i{=}1,\,U_t^{-i}{=}0\mid h_{1:t-1})
  =\int\! p(x\mid h_{1:t-1})\prod_j p(U_t^j{=}\delta_{ij}\mid x_j)\,dx
  =\eprd^i\prod_{j\neq i}(1-\eprd^j).
\]
Substituting into \eqref{eq:bayesnorm} ($c_d(\psi)$ cancels) and dividing
by $\prod_l(1-\eprd^l)$:
\[
  \gamma_t^i
  =\frac{r_t^i\,\phi_t^i}{\sum_j r_t^j\,\phi_t^j},
  \qquad r_t^i=\frac{\eprd^i}{1-\eprd^i}.
\]

\subsection{The affine per-chunk likelihood}

By the law of total probability, the likelihood of the query as a function
of one chunk's utility is
\[
  p(q_t\mid x_i,\,h_{1:t-1})
  \propto x_i\,\phi_t^i+(1-x_i)\,\phi_t^{-i},
\]
an affine mixture with hit weight $\phi_t^i$ (chunk $i$ used) and leak weight
$\phi_t^{-i}=\sum_{j\neq i}r_t^j\,\phi_t^j$ (some other chunk used).

\subsection{The utility posterior and its projection}
This update is not used in the experiments; it is included for completeness. Fix chunk $i$ and abbreviate $e:=\eprd^i$, $n:=\nprd^i$. The posterior
density is
\[
  p(x_i\mid q_t,h_{1:t-1})\propto\Bet(x_i;e,n)\,[x_i\,\phi_t^i+(1-x_i)\,\phi_t^{-i}].
\]
Using the identities
$x\,\Bet(x;e,n)=e\,\Bet(x;\tfrac{en+1}{n+1},n{+}1)$
and
$(1{-}x)\,\Bet(x;e,n)=(1{-}e)\,\Bet(x;\tfrac{en}{n+1},n{+}1)$,
the posterior is a two-component Beta mixture:
\begin{equation}\label{eq:mixture}
  p(x_i\mid q_t,h_{1:t-1})
  =w\,\Bet\!\left(x_i;\tfrac{en+1}{n+1},\,n{+}1\right)
  +(1{-}w)\,\Bet\!\left(x_i;\tfrac{en}{n+1},\,n{+}1\right),
\end{equation}
with $w=\frac{e\,\phi_t^i}{e\,\phi_t^i+(1-e)\,\phi_t^{-i}}=\gamma_t^i$
(the responsibility). The first two moments are
\[
  \E[X]=\frac{en+w}{n+1},\qquad
  \E[X^2]=\frac{(en+1)(en+2w)}{(n+1)(n+2)}.
\]
The projected belief $\Bet(\epst^i,\npst^i)$ matches these moments.
The mean gives the mean update directly: $\epst^i=(en+w)/(n+1)$.
The concentration follows from inverting the Beta variance identity
$\Var(X)=\epst^i(1-\epst^i)/(\npst^i+1)$:
\[
  \npst^i=\frac{\epst^i(1-\epst^i)}{v_t^i}-1,\qquad
  v_t^i=\E[X^2]-(\E[X])^2.
\]
Writing $\npst^i=\tau_t^i(\nprd^i+1)$ gives the tempering factor
$\tau_t^i$. For a hard use ($w\in\{0,1\}$) the
mixture collapses to one component and $\tau_t^i=1$: the standard
conjugate update $\npst^i=\nprd^i+1$.

\emph{The tempering bound $0<\tau_t^i\le 1$.} By the law of total variance,
\[
  v_t^i=\frac{\epst^i(1-\epst^i)}{n+2}+\frac{w(1-w)}{(n+1)(n+2)}.
\]
The second term is $\ge 0$, so $\npst^i+1\le n+2$, hence $\tau_t^i\le 1$
with equality iff $w(1-w)=0$ (hard label).

\subsection{Predict step: the Laplace bridge}\label{app:predict}

The dynamics \eqref{eq:ou} evolve the logit
$L_t^i=\logit(X_t^i)$ as a Gaussian OU process, but the filter
maintains beliefs in the Beta family $\Bet(e,n)$. The
\emph{Laplace bridge} \citep{westharrison,minka} is the assumed-density
step that maps between the two: given a Beta belief
$(e_{t-1|t-1},n_{t-1|t-1})$, we (i)~map to the logit-normal by
matching the mean and variance of $\logit(X)$ under the Beta, (ii)~apply
the linear Gaussian OU predict in logit space, and (iii)~map the
resulting logit-normal back to a Beta by moment matching.

\paragraph{Beta to logit-normal (step i).} For $X\sim\Bet(e,n)$, the
mean and variance of $L=\logit(X)$ are approximated by
\[
  \E[L]\approx\logit(e),\qquad
  \Var(L)\approx\frac{1}{e(1-e)\,(n+1)},
\]
using a delta-method (second-order Taylor) expansion of $\logit(\cdot)$
around $e$. This is exact in the large-$n$ limit and accurate for $n\gtrsim 2$.

\paragraph{OU predict in logit space (step ii).} The OU step
\eqref{eq:ou} with retention $a$, equilibrium mean
$\bar\ell=\logit(\bar e)$, and equilibrium variance
$\bar v=1/(\bar e(1-\bar e)\bar n)$ gives
\[
  \E[L_t]\!=\!a\,\logit(e_{t-1|t-1})+(1{-}a)\,\bar\ell,\qquad
  \Var(L_t)\!=\!a^2\,\Var(L_{t-1})+({1{-}a^2})\,\bar v.
\]

\paragraph{Logit-normal back to Beta (step iii).} Inverting the
delta-method approximation on the predicted logit-normal:
\[
  \eprd^i=\mathrm{logistic}(\E[L_t])
  \approx a\,e_{t-1|t-1}^i+(1{-}a)\,\bar e,
\]
where the approximation
$\mathrm{logistic}(a\,\logit(e)+(1{-}a)\,\logit(\bar e))
\approx a\,e+(1{-}a)\,\bar e$ replaces the geometric interpolation
of odds with a linear interpolation of means (accurate when $e$ and
$\bar e$ are not too far apart), and
\[
  \frac{1}{\nprd^i}=\eprd^i(1{-}\eprd^i)\,\Var(L_t)
  \approx\frac{a^2}{n_{t-1|t-1}^i}+\frac{1{-}a^2}{\bar n},
\]
recovering the predict recursion \eqref{eq:predict}. In practice we
adopt \eqref{eq:predict} as the \emph{definition} of the $(e,n)$
dynamics; the logit-OU process is the continuous-time motivation that
this recursion approximates via the delta-method round-trip (sometimes
called the Laplace bridge \citep{westharrison,minka}, though that term
more commonly refers to the Dirichlet--Gaussian map).

\section{Derivation of the key fit}\label{app:key}

The key adaptation maximizes the conditional likelihood of the queries
given the labels. For labeled events where chunk $i$ is marked used,
the log-likelihood of the query under the vMF emission is
\[
  \ell(\{k^{i,c},\omega^{i,c}\})
  =\sum_{s\in\mathcal{C}_i\cup\{0\}}
    \log\Big(\sum_c\omega^{i,c}\,e^{\psi\,q_s^{\top}k^{i,c}}\Big),
\]
where $\mathcal{C}_i$ is the set of labeled events for chunk $i$ and
$q_0=k_0^i$ represents $n_p$ copies of the base embedding included as a
prior to anchor the keys near their initial position.
This is the standard vMF mixture log-likelihood \citep{banerjee2005}.
The E-step computes within-chunk responsibilities $\zeta_s^{i,c}$
(\eqref{eq:zeta}). The M-step updates each key as the normalized
responsibility-weighted resultant (\eqref{eq:keybatch}) and each weight
as the responsibility mean.

\subsection{When two keys separate}\label{app:saddle}

Take $C_i=2$ with equal weights and keys $k\pm\delta$, $\delta$ small
and tangential to $k$. For a query $q$ with tangential component $u$, the
within-chunk responsibility is
$\zeta^{1}=\mathrm{logistic}(2\psi\,q^{\top}\delta)\approx\tfrac12+\psi\,u^{\top}\delta$.
The M-step resultant for key 1 is
$\sum_s\zeta_s^{1}q_s\approx\tfrac{M}{2}\rho\,\bar k+\psi\,S\,\delta$,
where $M$ is the number of labeled queries, $\rho\bar k$ their mean, and
$S=\sum_s u_su_s^{\top}=M\Sigma$ the tangential scatter. Normalizing by
the resultant length $\tfrac{M}{2}\rho$ gives the new offset
$\delta'\approx(\psi/\rho)\,\Sigma\,\delta$, and key 2 receives $-\delta'$.
The fused point is therefore unstable in the direction of the top
eigenvector of $\Sigma$ iff $\psi\,\mu_{\max}/\rho>1$; for concentrated
query clusters $\rho\approx1$, giving \S\ref{sec:adapt}'s
$\psi\,\mu_{\max}>1$. Averaging (the $\rho$ in the denominator) pulls the
keys together; specialization (the $\psi\Sigma$ term) pushes them apart.

\section{Checking the approximations} 
The inference of §3 uses three approximations: a posterior factorized across chunks, a Beta projec-
tion, and the Laplace bridge between the logit-Gaussian of the OU dynamics and the Beta belief.
To measure their cost we simulated data from BRAG’s own generative model (30 or 50 chunks,
two keys each, 200 steps, 5 seeds, 55 runs) and compared BRAG with a joint particle filter over all
chunks, with 65.5K particles (doubling to 131K moved posterior means by less than 0.003). Because
the exact posterior is calibrated under the model, the distance between BRAG’s belief and that pos-
terior measures both the approximation error and the loss of calibration. BRAG’s beliefs stay within
0.02–0.09 of the exact posterior. Factorizing across chunks adds a mean error of 0.002–0.007, at the
particle filter’s own noise floor; the Beta projection adds 0.02 nats of KL divergence and the Laplace
bridge another 0.01, with top-3 agreement with the exact filter of 0.975 (0.988 without the bridge).
Recall@3 stays within ±0.015 of the exact filter. Taking the maximum over keys or averaging over
them gives the same Recall@3 ($-0.004\pm0.008$); on PersonaMem-v2, replacing the nearest key
with the full mixture changes Recall@5 by $-0.003$ [$-0.004$, $-0.001$].

\end{document}